# Machine Learning for Mediation in Armed Conflicts

Dr Miguel Arana-Catania[1]
Dr Felix-Anselm van Lier[2]
Professor Rob Procter[1*]

[1]University of Warwick and Alan Turing Institute
[2]Max Planck Institute for Social Anthropology

*Corresponding author: rob.procter@warwick.ac.uk

**Abbreviations**: NLP: Natural Language Processing; BERT: Bidirectional Encoder Representations from Transformers; LDA: Latent Dirichlet Allocation; NMF: Non-Negative Matrix Factorisation;

Abstract
Today's conflicts are becoming increasingly complex, fluid and fragmented, often involving a host of national and international actors with multiple and often divergent interests. This development poses significant challenges for conflict mediation, as mediators struggle to make sense of conflict dynamics, such as the range of conflict parties and the evolution of their political positions, the distinction between relevant and less relevant actors in peace making, or the identification of key conflict issues and their interdependence. International peace efforts appear increasingly ill-equipped to successfully address these challenges. While technology is being increasingly used in a range of conflict related fields, such as conflict predicting or information gathering, less attention has been given to how technology can contribute to conflict mediation. This case study is the first to apply state-of-the-art machine learning technologies to data from an ongoing mediation process. Using dialogue transcripts from peace negotiations in Yemen, this study shows how machine-learning tools can effectively support international mediators by managing knowledge and offering additional conflict analysis tools to assess complex information. Apart from illustrating the potential of machine learning tools in conflict mediation, the paper also emphasises the importance of interdisciplinary and participatory research design for the development of context-sensitive and targeted tools and to ensure meaningful and responsible implementation.

---

Policy Significance Statement
This study offers insights into how machine learning tools can be used to assist conflict mediators in organising and analysing data from highly complex and dynamic conflict situations. Machine learning tools can bring significant efficiency improvements to mediation by organising complex data and making it more easily accessible, giving mediators more control over existing information. They can also support moves towards consensus by highlighting areas in which political actors are converging or diverging; point to potentially overlooked areas of conflict or dialogue bottlenecks; and challenge prejudices that may have built up during a mediation process. This study offers a concrete example of how innovative machine learning tools can be used to address mediation in complex, fluid and protracted conflicts.

---

## 1. Introduction

As digital technologies advance, their fields of application continue to grow  and their use in conflict resolution is no exception. Peace technology is increasingly used in a range of conflict related fields, ranging from conflict forecasting and the design of early warning systems (Perry, 2013); to the systematic documentation of human rights violations (TheWhistle, 2019; Uwazi, 2020); to fostering broader inclusion in peace making efforts (Hilbinger, 2020). In a recent report, the United Nations Secretary-General recognised the importance of technology for the UN's peace-making efforts, emphasising that 'engagement with new technologies is necessary for preserving the values of the UN Charter and the implementation of existing UN mandates' (United Nations Secretary-General, 2018, p. 4).

Conflict mediation, the "active search for a negotiated settlement to an international or intrastate conflict by an impartial third party" (Berridge & Lloyd, 2012, p. 239), is also witnessing a rapid increase of the use of digital technologies to enhance peace making processes. Key mediation institutions are starting to explore the opportunities technology may offer in the context of peace making. Among other initiatives, in 2018 the UN Department of Political Affairs has launched a "CyberMediation Initiative", together with other leading mediation organisations such as swisspeace, the Centre for Humanitarian Dialogue and the DiploFoundation, to explore "how digital technology is impacting the work of mediators in preventing and resolving violent conflicts world wide" (Centre for Humanitarian Dialogue, 2018).

The search for new tools to support mediation efforts is at least partly motivated by the shifting character of conflicts in which mediators operate today. While the number of traditional symmetrical conflicts fought between states is declining (i.e., between armies), there is an increase of intrastate violence and asymmetric wars (i.e., between state armies and non-state actors), including civil wars, insurgencies, terrorism, guerrilla wars and large-scale protest and violence. Conflicts tend to be more fluid and fragmented, involve increasingly complex webs of state and non-state actors, and often spread across borders and affect broader regions (Strand et al., 2019). To a degree, the complexity of today's conflicts is fuelled by the near-ubiquitous spread of communication technology (Berman et al., 2020). Its low-barrier access means that information is available much more quickly to many more actors. This increases the number of actors that may engage in a conflict and "the rapid relaying of unverified information on, for example, a ceasefire violation, can raise expectations of a response and contribute to conflict escalation" (United Nations Secretary-General, 2017, p. 5). In addition, information leaks and other breaches of confidentiality, misinformation and disinformation, may hamper the mediators' peace- and trust-building efforts (United Nations Department of Political and Peacebuilding Affairs & Centre for Humanitarian Dialogue, 2019, p. 8). This shift in the nature of conflict means that violent conflicts have become more dynamic, complex and protracted.

These new empirical realities pose significant challenges to the international community's peace efforts and traditional forms of international peace interventions appear increasingly ill-equipped to address current conflict environments (Avis, 2019, p. 4). The heightened dynamic and complexity of conflict situations makes it more and more difficult for mediators to find avenues for peaceful political settlement of conflicts. In particular, mediators struggle to generate a solid understanding on how a conflict's context affects dialogue dynamics, including the range of conflict parties and the development of their political positions, the distinction between relevant and less relevant actors for peace making, or the identification of key conflict areas and their interdependence.

So far, data analytics technologies and techniques are still only used to "a lesser extent in the context of ongoing mediation efforts; mediation practitioners have expressed much interest in understanding their potential" (United Nations Department of Political and Peacebuilding Affairs & Centre for Humanitarian Dialogue, 2019, p. 12). In particular, while there is growing appetite among peace mediators for AI tools and international organisations are already deploying AI methodologies to inform operations, "it is yet to be determined what AI-based tools can contribute, substantially, to mediators' understanding of conflicts" (Lindström, 2020). Several recent reports and studies explore the potential of technology for mediation (Betz et al., 2018; Höne, 2019; Jenny et al., 2018; Lippi & Torroni, 2015; United Nations Department of Political and Peacebuilding Affairs & Centre for Humanitarian Dialogue, 2019), but there is still "a lack of concrete examples and discussions that could bring discussions forward" (Mediating Machines, 2020). This article seeks to contribute to efforts to close this gap by presenting the results of a project that was designed to explore and develop machine-learning tools to assist an international mediator to analyse large-scale peace dialogue transcripts to: (1) better understand the positions of the different parties; (2) how the different party positions had evolved over time; and (3) whether dialogue efforts had created any "zones of possible agreement" which could be exploited in subsequent peace negotiations.

## 2. Case Study Context

Since late 2014, Yemen has been subject to a protracted regionalised war comprising multiple armed and political conflicts, which has eroded central government institutions and fragmented the nation into several power centres. The roots of the war stem from a failed political transition that was supposed to bring stability in the aftermath of an Arab Spring uprising that brought down Yemen's long-time president Ali Abdullah Saleh.

Since the outbreak of the civil war, the United Nations, the Office of the Special Envoy of the Secretary-General for Yemen (OSESGY) and other international actors launched multiple attempts to facilitate talks between conflict parties to reach political agreement on the conflict issues (Palik & Rustad, 2019). However, since the beginning of international peace efforts, the conflict has significantly grown in complexity. While the conflict is often portrayed as a conflict between two main forces, these two broad coalitions are increasingly fractured and loyalties are fluid. A multitude of actors, both national and international, are now directly or indirectly engaged in the conflict, motivated by various divergent goals. The political and military situation on the ground remains highly dynamic, parties show signs of internal divisions and are becoming increasingly intransigent and party positions continue to shift (International Crisis Group, 2020).

This context complicates the mediator's goal to find consensus and agreement for political settlement. As party positions are volatile and increasingly difficult to track, the perceived effectiveness of dialogue efforts suffered and it became increasingly unclear whether any progress was made. This context also affected the work of the organisation for which this study was done. The organisation had been providing support to OSESGY's efforts to reach a peace agreement by acting as mediators since 2016 in an ongoing dialogue between the parties. The primary goal of this study was to find ways in which machine learning tools could make the data collected over the years of dialogue more accessible and navigable to the mediators and to find new ways to assist them in the analysis of the data. This included, for example, finding ways to help the mediators to identify how party positions had evolved over time, i.e., whether parties had moved closer towards a consensus on particular dialogue issues or not.

### 3. Methodology Overview

Knowing and understanding the social and political context of their application is vital for developing digital tools that are both effective and ethical. This is particularly important when machine learning tools are employed on highly sensitive subject matter such as conflict mediation. The project was designed in an interdisciplinary manner, involving both data scientists and legal anthropologists. This approach was applied throughout the project, starting with the in-depth, empirical analysis of the case at hand, which provided contextual information informing the development of machine-based analysis of the parties' dialogues; to the development of machine-learning tools and the interpretation of the results they produced; to considerations about how data outcomes could be presented in a meaningful way to mediators.

The project followed a participatory design approach (Bodker et al., 1995). Since its emergence in the 1980s, participatory design has become acknowledged as being key to successful IT projects (Voss et al., 2009). Arguably, following a participatory design approach has become especially important now, however, as machine learning techniques that are still so unfamiliar to many, are now being widely applied to new products and services (Slota, 2020; Wolf, 2020). This unfamiliarity can then lead to unrealistic expectations of their capabilities (Tohka & van Gils, 2021). Project progress was discussed in bi-weekly meetings with the conflict mediation team. Initially, these focused on building common ground between the participants: (a) familiarising the mediators with NLP and machine learning concepts, techniques, capabilities and limitations; (b) enabling us to gain an understanding of the dialogue process, the data that it generated and how; and (c) then establishing some initial requirements for how NLP tools could support the dialogue process. As the project progressed, we were then able to present a series of prototypes to the mediation team for their comments and feedback. This enabled the requirements of the tools to be progressively refined around an evolving set of agreed use cases that captured more specific ways in which the tools could be applied in support of the dialogue process and conflict mediation.

Computational Grounded Theory is a rapidly developing field and a number of methods have now emerged (e.g., Nelson, 2017). However, from our initial discussions with the mediation team, it became clear that it would be important not to try to follow any particular method but to let the method - and hence the tools - to be driven by our discussions with them . This eventually led to the development of: (1) two information extraction tools to organise dialogue text into predefined categories and to derive latent issues from the dialogues text; and (2) data analysis tools to measure party distances (i.e, how close or how far apart the positions of the parties are) on specific topics. Although each of the tools independently provides particular insights into the dialogue data, more holistic and meaningful insights into dialogue dynamics can be extracted when they are used in combination. In the following, we will describe how the tools were developed but also how the results of each tool will need to be triangulated with other sources of data, both qualitative and quantitative, to validate the output and to reach meaningful conclusions.

The specific technical details of the methodology used are described in sections 4, 5, and 6.

## 4.    Data

The data for the project was mainly sourced from the dialogue sessions and expert meetings between Yemen's main political groups that the mediation team had conducted between 2018 and 2019. Each dialogue session was documented through "rough notes". These notes were not verbatim transcripts of dialogue sessions but relatively well- and consistently structured documents that captured dialogue between parties in an abbreviated form. Each note covered working sessions held between 2018 (6 sessions) and 2019 (8 sessions) and contains around 12,000 words for each session. All the notes together add up to 177,789 words. These rough notes served as the basis for the dataset which was used to develop the machine-learning tools.

In addition to this dataset we also had access to a number of internal documents, including over 30 detailed meeting reports, as well as the organisation's own systematic "comprehensive analysis" of dialogue developments. These documents provided important contextual information that helped guide and structure the development of the tools.

### 4.1.   Data Preparation Tool and Pre-processing Methodology

Before any tool development for the analysis of this data could be undertaken, the notes had to be cleaned and pre-processed and transformed into a structured dataset for the analysis. To this end, we produced a tool for data pre-processing and cleaning, which is capable of processing both existing data and any future notes provided they are formatted in the same way.[1]

The pre-processing of the texts includes: deleting non informative, unicode characters from the original word documents, identifying indentation format of the dialogues in order to build the conversation threads correctly, deleting non conversational text, correcting and uniformising entities spelling, detecting text shared by several entities, and abbreviation expansion. The preprocessing is done with the help of the NLTK[2] package and the Pandas[3] library.

The tool then automatically extracts and organises data from the notes into an csv file, organising the dialogues comments in the following fields: text, original raw file name, year, month, participant name, participant organisation, and participant multi-organisation (in case several parties are sharing a statement).

This dataset provided the basis for filtering and extracting more nuanced information via machine learning tools at subsequent stages of the process. In addition, the Data Preparation Tool generated a csv file that can be used by the mediators to carry out a simple information analysis or to retrieve specific parts of the dialogues. For example, it is possible to obtain the comments made by a specific party or a subset of parties within a certain time interval.

## 5.    Issue Extraction Tools

Following initial discussions with the mediation team, it was determined that the basic requirements that of these tools should satisfy would be to: (1) categorise the dialogue texts into a set of issues predefined by the mediators; and (2) identify latent issues that emerged from the dialogue text without manually pre-defining them. To do so, we proceeded with two different approaches: query-driven topic modelling and topic modelling.

### 5.1.   Predefined Issue Extraction: Query driven topic modelling

After years of facilitating dialogue, the mediators wanted to systematise all available notes and recordings into a "comprehensive analysis" of the dialogue sessions. This "comprehensive analysis", which entailed a list of what the mediator perceived as key conflict issues, was to provide an overview of dialogue development, as well as to organise the dialogue sessions and to sketch out avenues on how to reach agreement on these issues. This list of issues is important as it reflects the organisers' first-hand knowledge of the dialogues, as well as their own particular vision and analysis of the situation. However, the drafting of the comprehensive analysis proved challenging as the mediation team had to manually extract information from the available data, which had proved to be both ineffective and costly. The goal of the predefined issue extraction tool was to match dialogue text to 18 predefined conflict categories or issues defined by the mediation team in their own analysis.

---

[1] We have produced a style guide to ensure that any future notes can be read and analysed and any automatic text analysis tool is able to extract as much information as possible in the most efficient way.

[2] https://www.nltk.org/

[3] https://pandas.pydata.org/

**Method**

To create the predefined issue extraction tool, we used techniques based on query-driven topic modelling, where keywords relevant to each issue are predefined. The tool uses word representations produced with a model trained on a large corpora of text to understand how words are used in relation to each other, and be able to detect words with similar meanings or relating to the same matters. In this case the technique uses this knowledge to detect words in the comments that are related to the predefined keywords for each issue. For example, for the query term "State Institutions", the tool will not only identify passages in the text which contain the exact word but also related words such as "ministry, district, government", etc., because these have similar meanings to the query term.

The procedure consisted of defining query words for each of the 18 issues and detecting in the dialogues terms whose vector representations (embeddings) were found within a certain distance of the queries. The comments of the dialogues containing the original keywords or near terms were categorised as belonging to the issue. It was possible to categorise a text with multiple issues. All the definitions of the keywords, and analysis of the results was done qualitatively in collaboration with the mediation team.

We first defined an initial set of keywords for each issue, and compared the near terms found in the texts by using two general types of non-contextual word representations, Word2Vec (Mikolov et al., 2013a, 2013b) embeddings produced by Google and GloVe embeddings (Pennington et al., 2014) produced by the NLP group at Stanford University[4]. The GloVe embeddings were found to produce more meaningful results. For the rest of the evaluation we used the pre-trained GloVe embeddings "glove.6b", using 6 billion tokens represented in a 300 dimensional space. The embeddings are trained in the Wikipedia 2014 dump and in the English Gigaword Fifth Edition corpora (Parker et al., 2011). This selection was done considering also the need of the tools to be light enough to run in common computers by the mediation team. Stop words from the texts were removed by using the NLTK package. No further pre-processing steps were performed beyond those described in section 4.1. The distance between terms was computed as the cosine similarity between the embeddings by using the Gensim[5] library (Rehurek and Sojka, 2010). The threshold distance was defined dynamically, starting from a common value but with the possibility of reducing the distance for each term in case of obtaining too many related words.

The fine tuning of the tool included, as a first step, adjustment of the parameters and threshold to identify related terms in the query. The only parameters involved in this technique are the similarity range and maximum number of words to consider. After examining the outcomes of other configurations we selected a minimum threshold for the cosine similarity of 0.4, with the possibility of increasing it up to 0.6 when finding more than 1000 similar words. Next, we concentrated on refining search terms. The search terms for each issue needed to be iteratively refined for the tool to be able to distinguish between and associate text relevant to different issues. For example, to identify relevant text for the issue "The South", search terms such as "reparations", "independence", and "autonomy" were used.

Once the list of search terms for each issue was consolidated, we further refined the search by experimenting with different types of queries. First, we tested the results when the tool performed a search for each term individually, which means that the tool would associate text identified by an individual search term with a particular issue. We also tested the results when all of the search terms were combined. This final combination of searches with all keywords for each issue was used to produce the final categorisation of comments into the predefined issues. The system produced as an output not only the list of comments categorised, but also the query expansion for each of the initial queries of each category with the new near terms found for any future refinement of it.

Overall, the query-driven topic modelling tool produced promising results in terms of identifying and categorising party comments into the different predefined issues.

---

[4] https://nlp.stanford.edu/projects/glove/
[5] https://radimrehurek.com/gensim

## Discussion

The results of the machine-learning analysis were presented in a csv file, which categorises each comment made by each participant into one (or more) predefined categories. The filtering function of any spreadsheet software allows mediators to filter comments according to different parameters, including a particular dialogue category but also according to a year or month in which comments have been made, and particular parties or dialogue participants involved. Such a tool allows for the effective navigation and systematic exploration of dialogue data and may prove to be useful, for example, to hold participating dialogue parties accountable for their positions.

The tool also allows for a meta-level analysis of the data. In addition to categorising pieces of text, the relevance of individual issues were measured by the number of words, which allows for insights into broader dialogue dynamics. By looking at the words-per-issue graph (Figure 1), we can identify the amount of debate that each issue generated over the course of a year and how this changes over time. The data shows that there has been a general increase of dialogue activity from 2018 to 2019, as the number of words per issue approximately doubled from one year to the next. Looking more closely into each issue, we can also identify how intensively individual issues were discussed over time. Here, we can observe that while some issues remain relatively stable in terms of how intensively they are being discussed, others tend to fluctuate. For example, while discussions concerning "Decentralisation/federalism", "Dispute resolution" and "National body" increased from 2018 to 2019, the issue of "Government of National Unity" - and to a lesser extent "Demobilisation", "Guarantees", and "Sequencing of negotiations" saw a decrease of discussion.

For the purposes of interpretation of the data, it is important to remember that comments are categorised into one category or another based on their word content. This means that the activity reflects what has been said, regardless of whether the debate was organised at the time to address that issue. Thus, a topic may be debate-heavy because of the mediator's thematic emphasis in the dialogue sessions, but they may also highlight the issues that produced the most engaged discussions among the participants.

### 5.2. Latent Issue Extraction: Topic Modelling

The second type of issue extraction was designed to automatically extract, identify and classify the most relevant issues raised by the participants throughout the dialogues. Rather structuring the dialogues into a set of predefined issues, the tool identifies the most relevant issues as they emerge organically from the discussions, based exclusively on the textual content of the dialogues. This second method of issue extraction offers a new perspective on dialogues' substantive focus and may point mediators to aspects they have not yet considered.

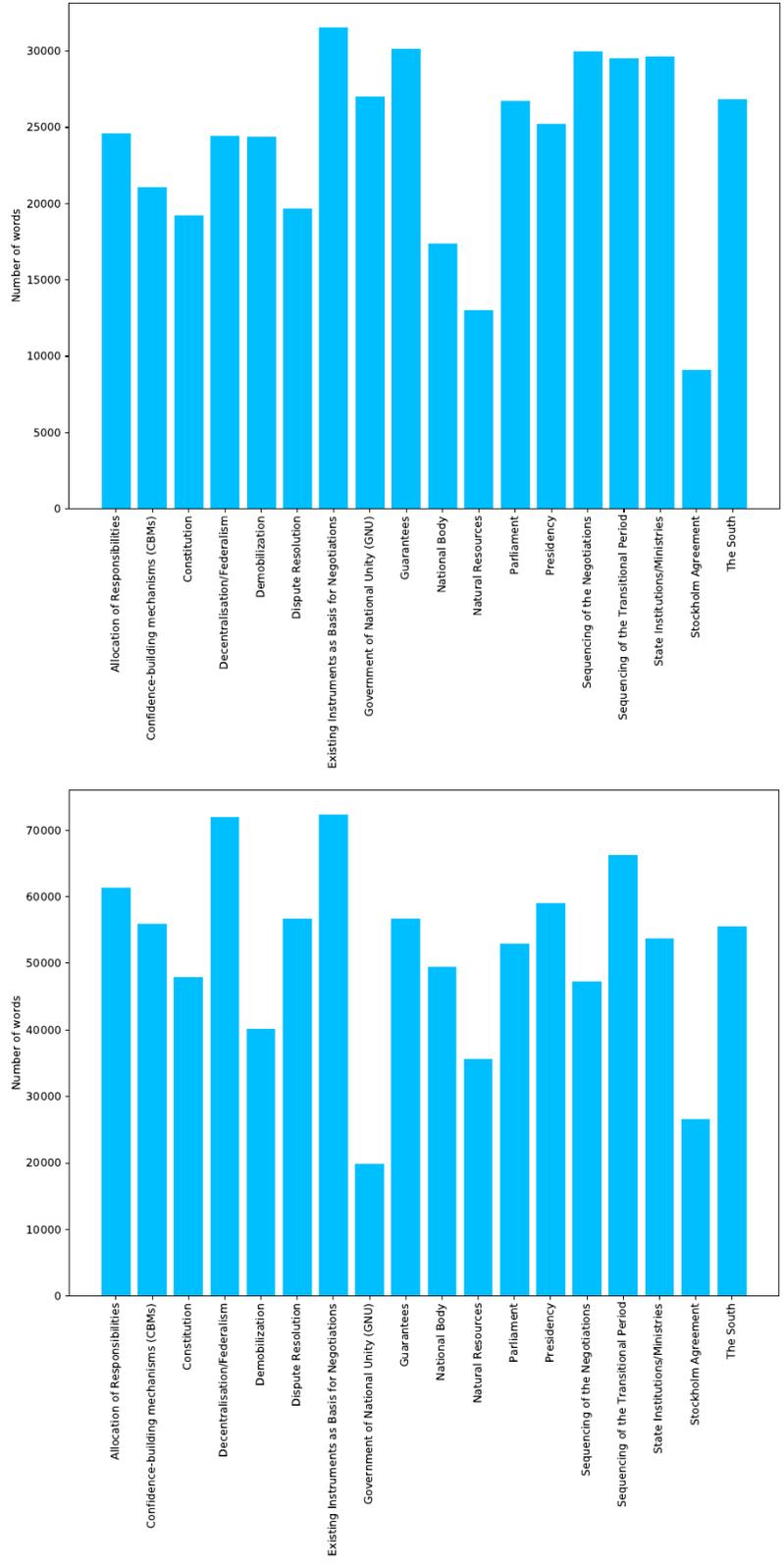

Figure 1: Relevance of predefined issue by number of words for 2018 (top) and 2019 (bottom).

## Methods

Two different topic modelling techniques, Latent Dirichlet Allocation (LDA) (Blei et al., 2003; Hoffman et al., 2010) and Non-negative Matrix Factorization (NMF) (Hoyer 2004; Lee and Seung 2001; Paatero and Tapper 1994) were tested and evaluated for the latent issue extraction. The underlying idea of both techniques is the same: given a set of documents in a corpus (in this case we define each comment made by a participant to be a distinct document in the dialogues corpus), topics consist of a set of words and each document is a mix in some proportion of one or more topics. NMF, iteratively searches for a matrix decomposition of the original matrix of documents into words. LDA, as described in its references, understands the documents as generated by probabilistic distributions over topics and words, where, in particular, the probability of each topic in each document is given by a multinomial Dirichlet distribution.

LDA and NMF were experimented with to automatically extract between 5 to 30 issues, each of which was then defined by 10 keywords that, based on their frequency of occurrence, were the most representative of the words that make up that issue. Together with the mediation team we then analysed the meaningfulness of the results by scrutinising the keywords of each issue as well as the top 10 extracted participant comments for each issue.

Following evaluation, it was determined that NMF produced better results. In particular we used the NMF implementation of the scikit-learn[6] package (Pedregosa et al., 2011), using the Frobenius Norm as the objective function, a Term-Frequency Inverse Document-Frequency (TF-IDF) representation of the words, Non-negative Double Singular Value Decomposition (Belford et al., 2011; Boutsidis and Gallopoulos 2008) for the initialization and a coordinate descent solver using Fast Hierarchical Alternating Least Squares (Cichocki and Phan 2009; Gillis 2014). Stop words from the texts were removed by using the NLTK package. For the vocabulary used in the topic modelling we considered the top 10000 features and excluded terms appearing in more than 90% of documents. Regarding the parameters of the NMF decomposition, we used a regularisation parameter alpha with a value of 0.1, and a regularisation ratio of 0.5 for the mixing between the L1 and L2 regularisation, and a tolerance of 1e-4 to the stopping condition. NMF outputs directly the decomposition of documents in each of the possible topics, without the need to identify the keywords in the documents and then identify the keywords in the topics. A 0.1 threshold was used as a percentage threshold to classify a document as belonging to a topic.

We then refined the topic modelling tool. For example, we tested whether the tool would perform better if it only considered nouns in the dataset. Further, we noticed that the structure of "working sessions" outcomes (i.e., multiple parties presenting results of group discussions on particular issues) tended to obscure the topic modelling process, as the substance of this text was often more technical in nature, making it more challenging for the tool to classify text meaningfully. We decided to drop multiple party responses from the overall analysis, which then produced the most well-defined issues and provided the richest forms of information. No other additional pre-processing steps were applied besides the ones mentioned here and in section 4.1.

## Discussion

The latent issue extraction tool allows for a more unconstrained analysis of the issues that emerge naturally from the dialogues. This approach may highlight aspects of the negotiations that appear to be of particular importance to the dialogue participants and may offer new perspectives on dialogue dynamics. While the distribution of the topics may still broadly reflect the mediator's choice of dialogue structure, the data reflects what participants actually said during the dialogues and how their interventions shaped what was being talked about at each moment. When comparing the issues generated through latent issue extraction with those predefined by the mediator we found some overlap. For example, questions about natural resources, the composition of the national body, the South, or sequencing were issue categories that appeared in the results of both extraction tools. However, the latent issue extraction tool also brought to light several "new" issues, such as representation, disarmament, the role of the UN, etc. While each of these issues may be categorised into one of the issues previously identified by the mediator, they nevertheless highlight aspects of the negotiations that appear to be of particular importance to the participants. This information could be used to reconsider the substantive focus of future dialogue sessions.

---

[6] https://scikit-learn.org

The tool produced a list of new issues and organised all relevant text into those issues (see Table 1 and Box 1 for an explanation).

| Issue | Issue Description | Keywords |
|---|---|---|
| issue 0 | Sequencing | political forces regarding military conflict security want war yemen sanaa |
| Issue 1 | Executive Powers/Composition | president powers vp presidential_council vps presidency advisory_council prime_minister decisions legitimacy |
| Issue 2 | Institutional Security Arrangements | would military_security_committee committee formed option implement third_party independent two monitor |
| Issue 3 | Framework Agreements | agreement hodeida implemented sides implementation framework interpretation_committee implement signing stockholm_agreement |
| Issue 4 | Implementation Issues | need address new solution violations realistic transition take end_war participation |
| Issue 5 | Sanctions | sanctions said guarantees yemeni talking actors think implementation agreements |

Table 1: First 5 latent issues and top ten keywords for 2019.

| Issue | Issue Description | Keywords |
|---|---|---|
| Issue 0 | Representation/Appointments (e.g. technocratic/political) | political parties minister social competence arms achieve important real conditions |
| Issue 1 | Allocation of Responsibilities | government president presidency political_forces gcc_initiative parliament technocratic would forces decision_making |
| Issue 2 | Natural Resources | resources regions federal natural draft local revenues given model draft_constitution |
| Issue 3 | National Body | national_body committee composition guarantees implementation current oversee body disputes role |
| Issue 4 | South | south make issue southern since represent southern_issue groups yemen international |

| Issue 5 | Disarmament/Guarantees | state legitimacy arms even armed militias groups issue solution |
| --- | --- | --- |

Table 2: First 5 latent issues and top ten keywords for 2018.

Each issue is defined by the most frequent words that appear in the comments regarding the issue. These keywords are presented on the right side of Table 1.

Each comment can be tagged as belonging to several issues at the same time (e.g., somebody talking about "Sanctions" and "Agreements" in the same paragraph). In order to understand the issues better, the tool also extracts the top 10 most representative comments for each issue. That is, the comments that are more uniquely identified by the specific issue in comparison with others.

The list of representative comments, together with the list of keywords of each issue, help to better understand what is being discussed. Using this information, we produced a manual description of each issue, that can be seen on the left side of the previous tables.

Box 1: Explanation of Tables 1 and 2

This form of issue generation can be used in various ways. Most importantly, latent issues offer an alternative perspective of the way that the dialogues unfolded in practice. The relevance of issues by number of words graph (see Figure 2), shows the activity of parties in each of the latent issues. For example, the data indicates that questions surrounding the issue number 0 "sequencing" appear to have been discussed intensively in 2019. The disproportionate amount of text in this issue is also indicative of significant overlap of the broader issue 0 "sequencing" with other issues. This was confirmed by measuring the overlap between issues.

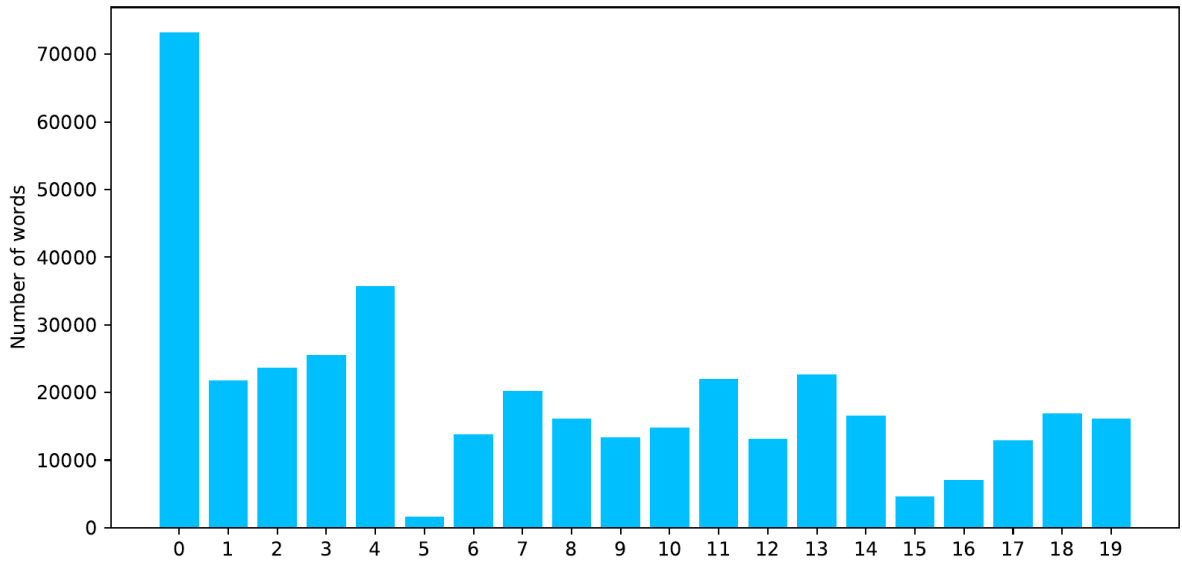

Figure 2: Relevance of issues by number of words, 2019 dialogues.

Another way of making use of latent issue extraction is to delve deeper into each issue by analysing the comments associated with it. This step allows for the most granular analysis of party positions by exploring what parties actually said on a particular issue.

The tool for latent issues extraction adds an additional column to the csv file containing all dialogue data to include the categorisation of each comment into the correspondent latent issue. This file can be used to filter the content related to specific issues, and also combining it with other previous filters as the ones allowing to select a specific party or time period.

## 6. Measuring Party Distances: Transformers Representations

The categorisation of text through issue extraction, both via the predefined issue extraction and the latent issue extraction, now provides a basis for the identification of party distances, that is, the substantive distance of one party position to another. The primary goal was to evaluate whether the mediation efforts had led to any convergence between parties. But it also allows for more detailed insights into which parties diverge on which issues and the behaviour of particular parties throughout the dialogue. This type of detailed analysis may be particularly useful in dynamic and complex conflict settings with multiple actors and multiple areas of conflict.

The categorisations of text via latent issue extraction and predefined issue extraction have allowed the extraction of what each party has said about each of the issues. To measure party distances, the two text categorisations were used to apply a technique that enables the evaluation of the proximity between texts. This textual distance is evaluated based on the words used and the context in which those words are used.

### Methods

To reach an understanding of party distances, we employed Bidirectional Encoder Representations from Transformers (BERT; Devlin et al., 2019), another language model for Natural Language Processing. This model is not based on a formal definition of language but is derived from developing a statistical understanding of how language is actually used. To achieve such an understanding and to produce a model for how language is constructed, the tool has been trained with large datasets (in this case the English Wikipedia dump and BookCorpus (Zhu et al., 2015) a dataset of 11,038 unpublished books). During this process, it assigns a "linguistic position" for each word in an abstract multidimensional space. Words with a similar meaning are assigned positions in close proximity in this space. The model can then produce insights about the relationship between two words by measuring the distance and direction between two words in this space. For example, the distance and position of the words "London" and "UK" is the parallel to the words "Berlin" and "Germany". In this case, the representation of the words is contextual; the position of each word is also defined considering the rest of the sentence in which the words has been used (e.g., the word "pupil" has a very different meaning when talking about "eyes" or when talking about "students").

This method is applied to the dialogue dataset to assess the distance between party positions. The assumption is that the model would be able to extract party differences by measuring the distance of the language used by one party from that of another party.

In this project we used the BERT implementation of HuggingFace[7] 'bert-base-uncased' with 12-layer, 768-hidden parameters, 12-heads, and 110 million total parameters. The pre-trained model used can be found in the previous link. Texts longer than the token limit were split in order to avoid truncation. For long texts the embedding representation was obtained as the mean of the representations of its components. The distance between terms was computed as the cosine similarity, by using the Gensim library.

### 6.1. Party Distances in Predefined Issues

We produced several graphs to illustrate party distances. In the first set of graphs, party distances are measured against an average linguistic position of all parties (Figures 3 and 4).

The top straight line represents the average position. This is not the position of a specific party but an average position of the four selected main parties. Having defined the position of each party in that "linguistic space" (calculated from the phrases used by them in the dialogues), we can also calculate in that same space what would be the average position between them, which would represent the position of consensus closest to all of them. The line of each party then represents each party's distance from the average position (the lower in the graph the farther from the average position).

---

[7] https://github.com/huggingface/transformers

The second graph (Figure 4) displays party distances against the position of a particular party, whose position has been chosen as a baseline.

Both graphs represent the average position of each party over one year of negotiations (2019). It is important to note that, at this level of abstraction and with the limited amount of data available, the graphs cannot display the fluctuation of party positions during one particular dialogue session.

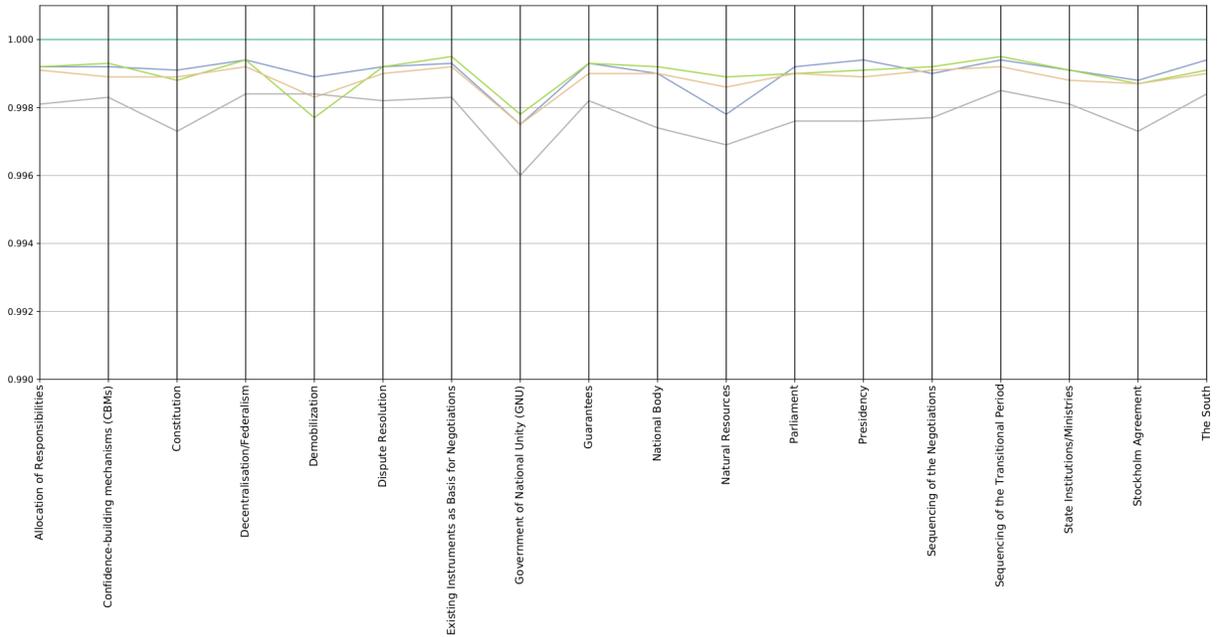

Figure 3: Party positions measured against average party position for 2019.

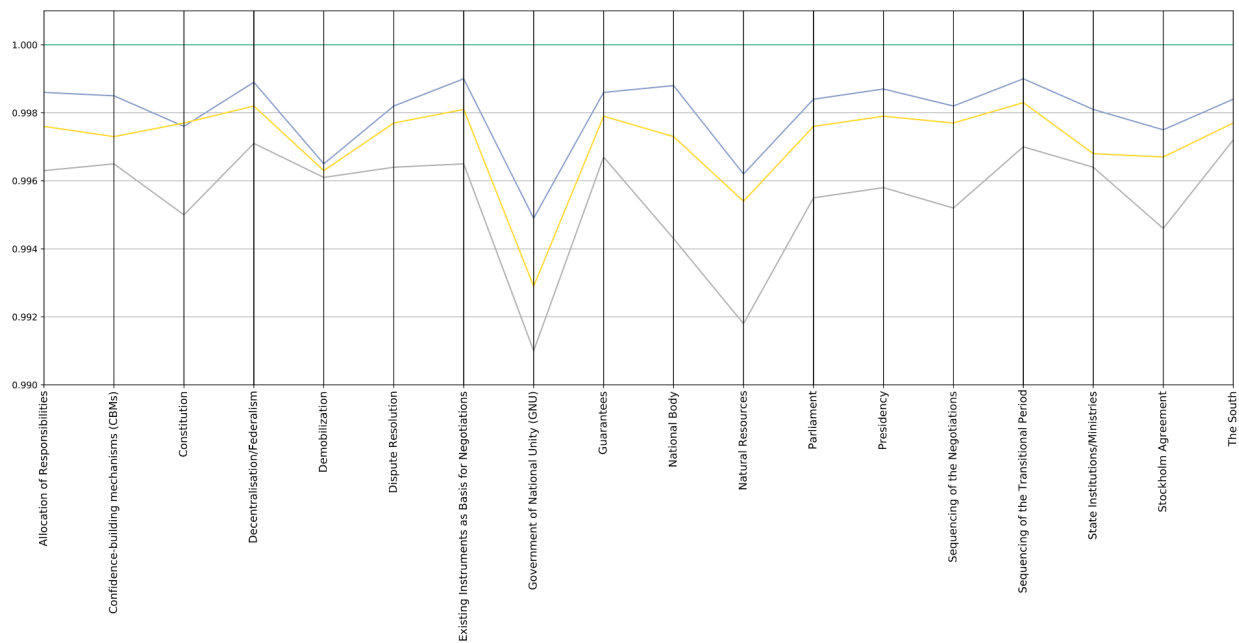

Figure 4: Party positions measured against a baseline position of a single party for 2019.

It is also important to clarify that in the two graphs each party represents its distance only from the reference position (average or baseline party). The fact that the lines of two parties cross does not mean that they are close to each other, it only indicates that both are at the same distance from the reference position. For the purpose of interpreting the graphs and the distances between parties it is useful to bring to mind the image of "parties in a room". If the centre of the room is the average position of all parties, then parties could be equidistant to the centre, but positioned at different corners of that room. Hence, parties that appear to be close on the graph could hold diametrically opposed political views. To inspect the specific distance between any two parties we need to refer to the next set of graphs (heatmaps), which show this more detailed information about each pair of parties.

It is further crucial to highlight that the limited amount of data available generates a margin of uncertainty in the graphs. It is difficult to obtain a precise value of this uncertainty, but we have observed that the size of the differences between parties is in the order of magnitude of the changes observed in any party position when varying the 10% of the text of the party. Thus, the variations should be considered within the uncertainty margins. This implies that the above differences between parties should not be taken literally but should only be used as indicators to serve as a prompt for internal reflection on the mediators' own perception of the dialogues. For example, when using the above graphs mediators should focus on relative trends between parties, and not on the absolute values of the differences.[8] Generally, charts should always be analysed in conjunction with the activity chart depicted in Figure 1. In cases where the amount of information is low, there is a considerable increase in the margins of uncertainty for the distances. This means that in those cases peaks observed in these issues in the last two graphs have a much higher probability of being produced by this lack of data, than by a difference between the positions of the parties.

The most significant dips in the graphs, such as the systematic positioning of one of the parties as the most distant from the reference positions, or the increase in distance of some issues, are the clues we get from these graphs that invite a more detailed internal analysis of these matters.

Heatmaps complement the analysis of party positions shown in the graphs as they allow for a closer examination of the distance of different parties to one another and can be used for a more focused analysis of party distances on particular issues. For example, the graph depicted in Figure 4 reveals a greater distance on the "Natural Resources" issue. This would prompt a closer analysis of this issue on the basis of the heatmap to explore more closely which parties diverge. Below we can see one of these heatmaps (Figure 5).

---

[8] For a more detailed discussion of the values in these graphs refer to the Appendix.

The parties appearing in the row and column of each tile are the same as those used in the comparison. Four levels of distances are considered, going from light green meaning smaller distance (that is why the diagonal presents this colour, since it shows the position of a party compared to itself) to red, meaning larger distance.

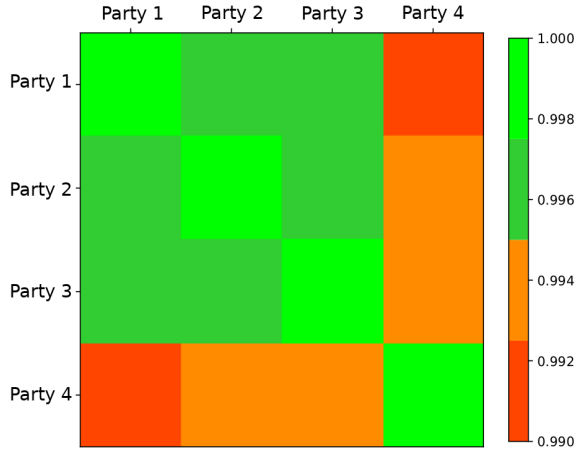

Figure 5: Party positions between each pair of the selected parties for the issue "Natural Resources" for 2019.

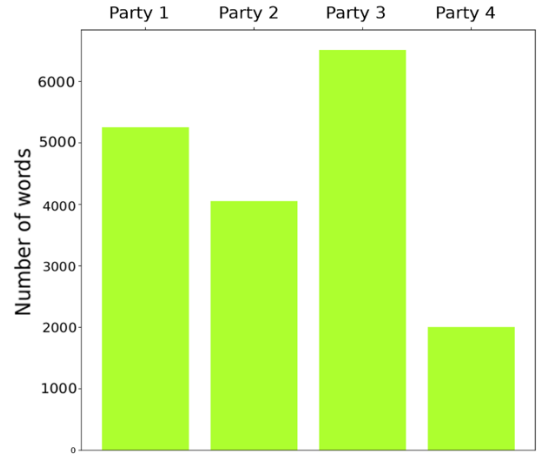

Figure 6: Number of words of each party for the issue "Natural resources" for 2019.

Again, as explained previously, it is important to emphasise the uncertainty in the data. To assess the viability of the data it is important to scrutinise the amount of available data not only for the whole issue but for each specific party when talking about the issue. To this end, we can make use of the party activity graph (Figure 6).

Besides helping to assess the graphs, the comparison between the activity of different parties on a specific issue could prove useful when it comes to understanding whether there are parties that are dominating dialogue sessions on particular issues. Such insights could support the strategic organisation of the dialogue sessions. For example, they could prompt dialogue organisers to re-consider the allocation of speaking time; or help to identify potential bottlenecks in negotiations and engage in bi-lateral dialogues if a particular issue appears to be of significant importance for a particular party.

### 6.2. Party Distances in Latent Issues

The same process was repeated with the results produced by the automatic generation of issues.

Next, we present graphs comparing the party distances on different issues. As explained in Section 5.2, in this case the issues on which the distance is evaluated have not been selected manually but generated automatically from the text. These issues are defined by a series of keywords listed in Table 1, where each one has been given a label based on its keywords and analysis of the representative comments on each issue.

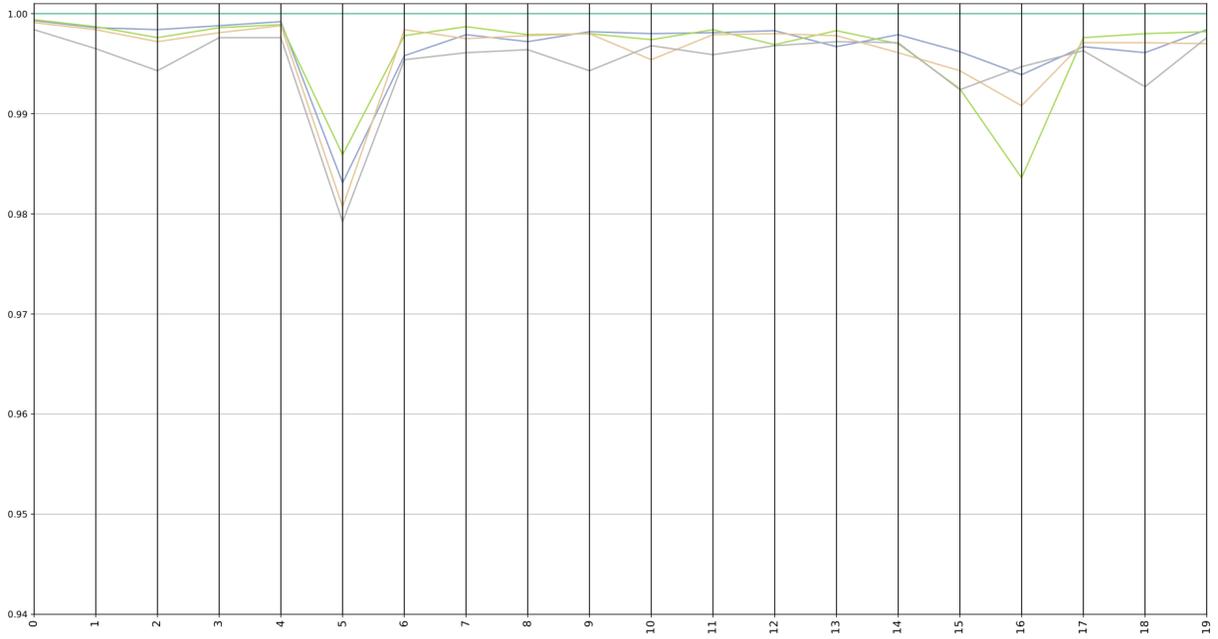

Figure 7: Party positions measured against average party position for 2019.

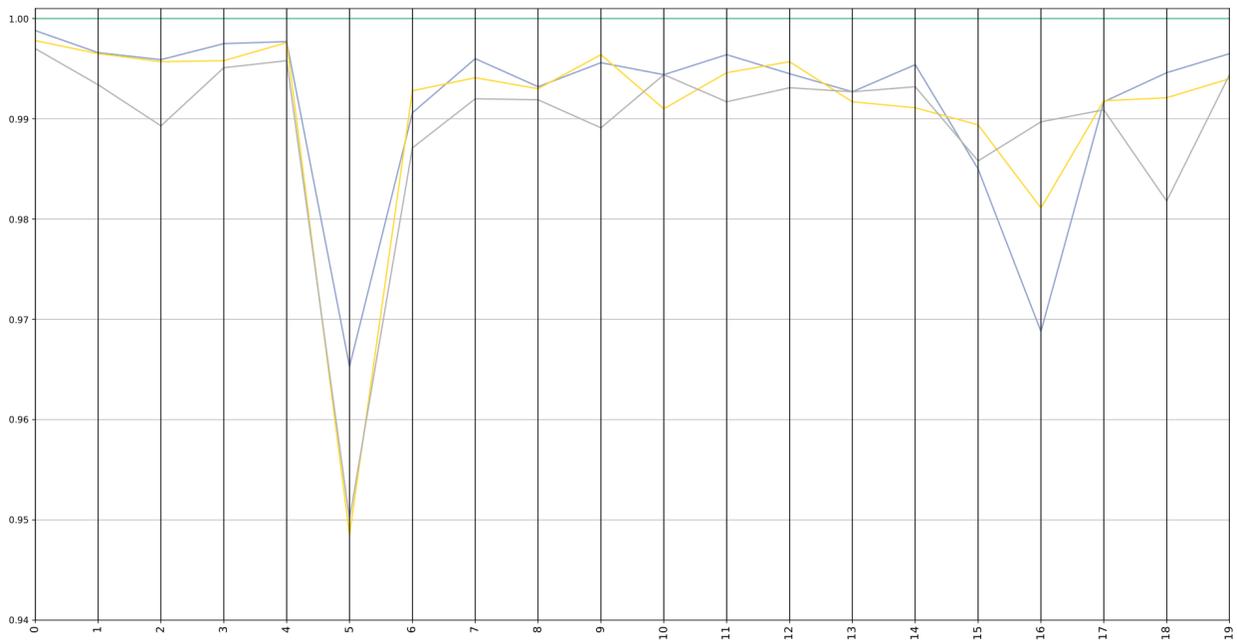

Figure 8: Party positions measured against average position of a baseline party for 2019.

As in the query driven case, the analysis was conducted by taking as a reference the average position of the parties and the position of a specific party. It is important to emphasise that the large dip in issue 5 is directly related to the lack of data of that issue as shown in Figure 2. For a more general discussion on these graphs refer to the previous query driven section.

Comparing both graphs with the predefined issues case it can be seen how the distances are reduced here. This may suggest that restructuring the conversations into the latent issues that emerge from the comments made by each party could make it easier to find consensus positions between the parties.

Below is an example of a heatmap (Figure 9) and its associated party activity graph. We have selected this issue from the previous graph as one of the issues where the largest differences are observed, which we observe again in this heatmap. However, we observe that in this example the amount of data is only 25% of the example in the previous heatmap. Hence the differences between party positions have larger uncertainty margins and should be considered less robust than issues that have more data.

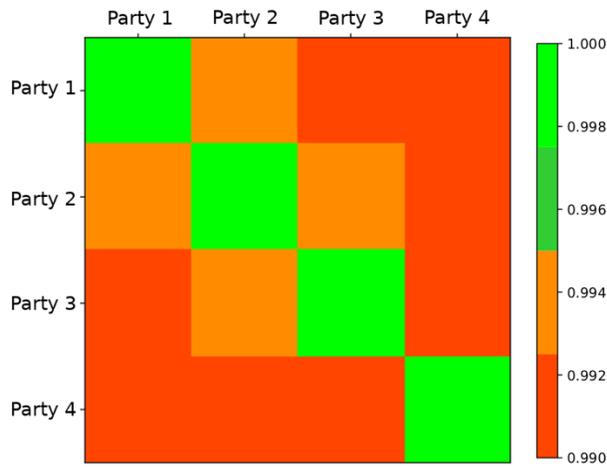

Figure 9: Party positions between each pair of the selected parties for issue 18 for 2019.

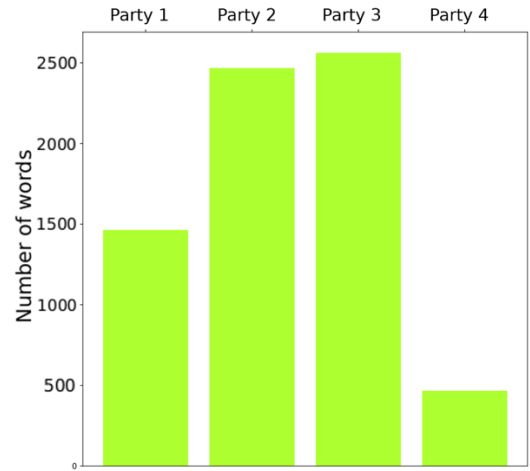

Figure 10: Number of words of each party for issue 18 for 2019.

## 7. Discussion

Overall, the study supports growing evidence that machine learning tools have the potential to provide meaningful insights into highly complex and dynamic mediation processes and to support consensus finding by providing tools for effective knowledge management and innovative dialogue analysis.

### 7.1. Knowledge Management

At the most basic level, machine learning tools can effectively support mediators in managing knowledge accrued over several years of peace dialogue. Often, as in the case at hand, valuable data may be available but is too complex and too vast to be assessed without technological support. The tools developed in this project have demonstrated that machine learning can bring significant efficiency to mediation by organising complex data and making it more easily accessible. The Predefined Extraction Tool, for example, has significantly eased the process of organising large amounts of dialogue text into predefined issues - a task that previously took several researchers weeks to accomplish. While machine-learning driven categorisation may not yet be as precise as a human-driven one, the tool still allowed the organisation to revise and update their "Comprehensive Analysis" more rapidly and conveniently. The tool also significantly improved the accessibility of available data. Presenting and organising data outputs in a csv file allows mediators to perform targeted searches by filtering information on the basis of a number of different variables. For example, mediators can now quickly obtain information on what a particular party had to say on a particular issue at any given time or track how the position of a party on an issue has changed over time. This could then be used to further investigate why shifts in a party's position may have occurred and to hold parties accountable to particular positions or to concentrate work on particularly controversial issues.

### 7.2. Dialogue Analysis

Beyond knowledge management, the tools serve as an additional analytical tool in the mediator's toolbox by helping them to better grasp conflict dynamics and party positions and to adapt their dialogue strategy accordingly. A deep understanding of a conflict situation, the key conflict areas and how they may be interrelated, its actors and their interests, the relationship between different actors, and their potential openness for finding alternative solutions is key for devising an effective dialogue strategy (Amon et al., 2018). Machine learning tools may be particularly useful in complex and lengthy mediation processes in which viewpoints may calcify and prevent consensus finding and help to challenge stereotypes and prejudices that have built up during a mediation process. The project has demonstrated how machine learning tools can offer new perspectives on dialogue dynamics, by highlighting potentially overlooked areas of conflict, "Zones of Possible Agreement", dialogue bottlenecks, or connections between different conflict areas. This sort of analysis can then support mediators to adapt their dialogue strategies.

For example, mediators could use the Latent Issue Extraction tool to identify potentially overlooked but pressing dialogue issues that emerge from the dialogue discussions. The Party Distances Tool could then be used to better understand which issues are leading to discrepancies and whether they relate to particular parties. In addition, mediators could use the party activity graph to detect whether particular parties are dominating dialogue sessions on specific issues. The mediator can then use this data as a prompt to engage in a targeted analysis of dialogue text on the respective issue. This process can then provide the basis to fine-tune dialogue proceedings by considering allocation of speaking time or to engage in bi-lateral negotiations to address issues that appear to be of importance to a particular party. Strategic dialogue focus can also be supported by using the analysis of overlapping issues to identify connections between different issues and to find ways of addressing them simultaneously.

More broadly, tools could be used to assess the impact of a mediators' dialogue efforts. Especially in complex and protracted mediation scenarios, progress is often notoriously hard to track. For example, the Party Distances Tool can be used to examine whether party positions have converged over time or whether they have stagnated or even further drifted apart; they could use the issue activity graphs to identify if the organisation of the dialogues is congruent with the issues that were prioritised in the dialogue activity of the parties, and how party activity on each issue fluctuates over the years; or identify the impact of external factors on the dialogue sessions by analysing how party positions have shifted at particular points of the dialogue proceedings. In this case, the analysis of party distances from 2018 to 2019 for both the predefined issues and the latent issues shows that, overall, party positions appear to be converging, although this should be confirmed via a detailed analysis of dialogue transcripts. Also, when analysing the number of words per issue graphs, the data shows a significant increase of dialogue activity from 2018 to 2019. Such analyses can prove useful for a meta-level assessment of the overall trajectory of peace negotiations and the effect of a mediator's work on consensus finding. The Latent Issue Extraction tool can also be used to analyse whether there are any new issues that emerge from one year to another, or whether discussion on others has receded; or to detect which parties are closer to the average positions and which parties are further away on each issue. This information can then be used to capitalise on emerging "Zones of Possible Agreement" by focussing on issues where party positions appear to be converging and create momentum for successful dialogue by finding agreement on these issues. Alternatively, the mediator could concentrate work on particularly conflictive issues to foster consensus-finding on dialogue aspects that are creating bottlenecks.

### 7.3. Challenges and Limitations

One of the key limitations of the study is the limited amount of data available for analysis. Data analysis for this project was conducted on "rough notes", rather than verbatim transcripts of dialogue sessions. This limitation of data volume is likely to affect data analysis as it has the potential to skew results and to misguide its interpretation. This is particularly evident in the case of measuring party distances. Here, limited data prevents a more detailed analysis of party positions over time and can obscure party differences and skew results, potentially leading to misguided interpretation. This problem is exacerbated when more detailed analysis is sought for shorter timeframes, as the amount of data will be further reduced. As mentioned, mediators will need to be conscious of these limitations when using the tool and merely use its outcomes as a prompt to corroborate them via deeper qualitative analyses of original dialogue text.

The project also highlighted persisting challenges of Natural Language Processing technology in identifying nuanced vocabularies and their context and dealing with word play and ambiguity (Höne, 2019, p. 12). These challenges were evident when seeking to distinguish between closely related issues that address overlapping subject matters. In the case at hand, for example, issue extraction tools struggled to separate related but independent issues dealing with institutional matters, such as the issue of "National Body" or "Government of National Unity" or with procedural questions of the transition, such as "Sequencing of Negotiations" and "Sequencing of the Transitional Period". Here, close collaboration with mediators who understand linguistic subtleties of a particular context was instrumental in refining the precision of machine learning tools. However, more work will need to be done to improve the precision of Natural Language Processing tools when dealing with linguistically and substantively closely related matters.

While some of these issues may be mitigated by providing larger amounts of data, e.g. by analysing verbatim dialogue transcriptions, they generally point at the necessity to invest significant time and resources on collaborative tool-development. Complex and sensitive mediation processes remain a human-centric trade and AI-tools can merely offer additional analytical support to devise mediation strategies. Responsible and human-centric use of AI-tools means that mediators will require data literacy, which provides them with a basic understanding of the technological underpinnings of the tools and a methodological framework (i.e. "from the data to the graphs and back to the data") that allows them to critically assess data outcomes, as well as to explain and justify their decisions. Data dashboards may help to present data outputs in more accessible ways and help guide responsible and informed interpretation. However, data literacy and ensuring "meaningful human control" (Höne, 2019, p. 11) over the implementation of digital tools is still best achieved when mediators are actively involved as collaborators throughout the development of the tool.

Consistent and close collaboration between mediators and data scientists will also help mediators to develop a clear understanding of the kind of problems they envision AI tools to address and limits the risk of the development of tools that are skewed by cognitive or social biases as well as a lack of knowledge of data scientists. Participatory project design will contribute to accurate, ethical and context-sensitive tool development that can substantially contribute to a mediators' understanding of the conflict.

## 8. Conclusions

While data analytics has been employed in broader humanitarian and peacebuilding contexts, e.g., for the purpose of conflict analysis, early warning, prediction of conflict, such tools have not been extensively used in the context of mediation efforts. The application of machine learning in this case study has shown that these tools can play a significant role in forming comprehensive conflict analyses and informing mediation strategies. Such tools become particularly pertinent in the context of the emergence of more dynamic and complex conflicts. This study demonstrated that machine learning tools can effectively support mediators in: (1) managing knowledge by analysing large amounts of unstructured data; and by (2) providing them with new analytical tools that may lead to new perspectives on a conflict scenario.

The project also emphasised that machine-learning tools cannot replace human analysis, particularly in highly sensitive contexts such as conflict mediation. Meaningful and responsible development and deployment of machine learning tools requires an interdisciplinary and participatory approach throughout to help develop an understanding among users of machine learning's capabilities and limitations and to help data scientists to create context-sensitive, targeted, effective, trustworthy and ethical tools.

We are currently awaiting further feedback from the conflict mediation team of their experiences of applying the tools and we will use this to further refine them. The introduction of new tools into work practices almost inevitably requires adaptation on the part of their users; this, in turn, leads to the need to adapt the tools, as users gain familiarity with the tools, discover their strengths and weaknesses and new requirements emerge: that is, what we have referred to elsewhere as 'design-in-use' (Hartswood et al., 2002a; Hartswood et al., 2002b; Bodker & Kyng, 2018). This, we argue, applies particularly to machine learning tools, such as the ones described here, that are intended to assist - and not eliminate - human analysis. Hence, we are continuing to work with the mediation team in order to learn from their experiences of using the tools and thereby help to evolve them in ways that are productive for conflict mediation.

**Appendix A: Structure of the word representation space.**

In this project we work with numerical representations of text dialogues, and it is therefore important to take into account the structure of these representations and how this structure can affect the analysis of these texts. We have applied different models to represent texts, but we will limit ourselves in this appendix to studying the case of GloVe, although the conclusions can be extended to other models such as BERT.

As mentioned previously, we have used the 'glove.6B' pre-trained embeddings of GloVe. This model has a vocabulary of 400,000 terms, each of which is represented by a vector with 300 components. If we analyse how these vectors are distributed in this 300-dimensional space, we can see that they are distributed anisotropically. In Figure A.1 we represent for each of the vocabulary words which is its largest component; in other words, towards which of these 300 directions each of the vectors is mainly oriented. As we can see, there are directions with a higher density of vectors than others. Figure A.2 shows the same information as a histogram, confirming the prevalence of some directions over others and therefore the anisotropy of the space of representations.

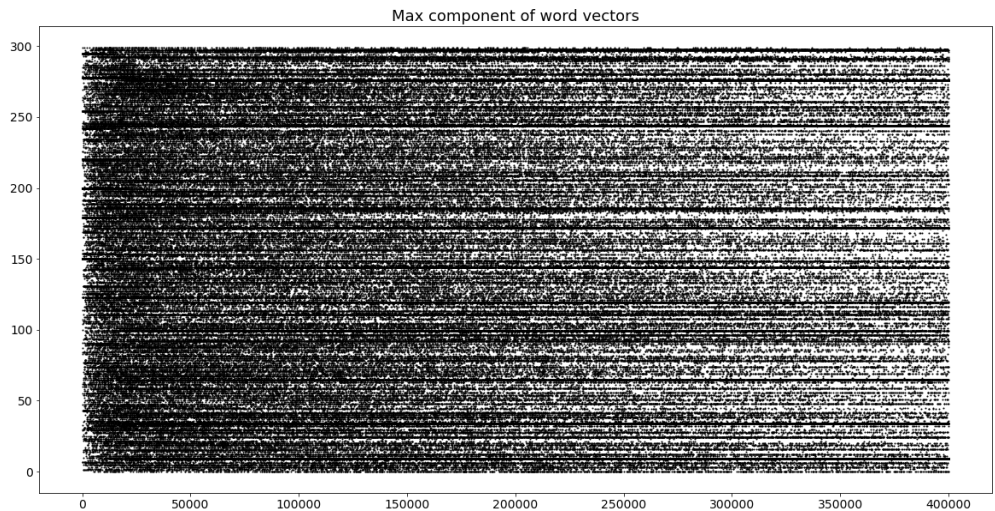

Figure A.1

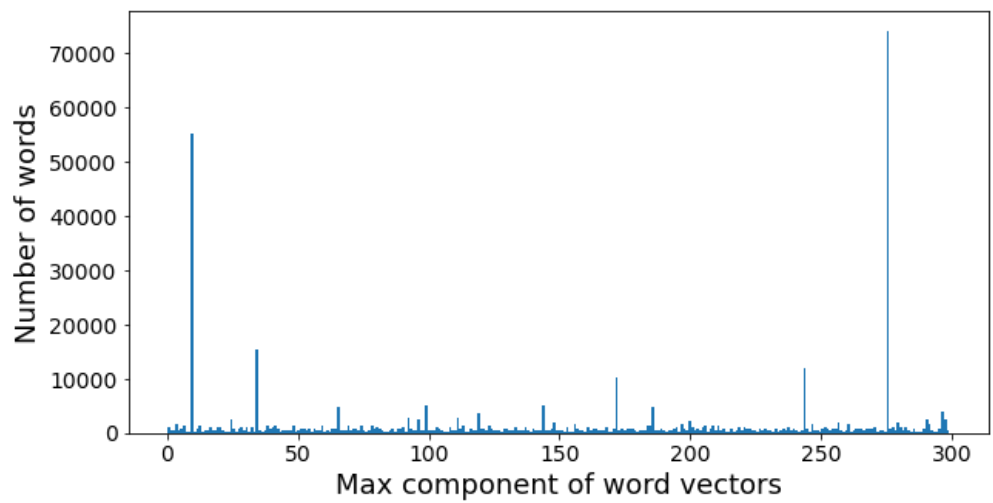

Figure A.2

In the following, we will show the effects of such a structure when operating on long texts. To show the universality of this analysis, we will take as an example of analysis two corpora which are very different from each other and also with regard to the texts used in this project: Corpus 1 = the text of 'The Wisdom of Father Brown' by G. K. Chesterton; Corpus 2 = the text of 'Leaves of Grass' by Walt Whitman.

In Figure A.3 we plot in the top row the value of the largest and smallest components of each of the first 5000 words of Corpus 1 (we include the smallest to take into account all possible orientations of the vectors). In the bottom row we represent the same quantities, but in this case for the average vectors up to each term; for the term number n we take the average of the first n vectors. As we can see in this row, although at the beginning the extreme values oscillate, when a sufficient number of words are considered, they converge towards a value with respect to which they remain approximately stable. In Figure A.4 we can see the same behaviour in the case of Corpus 2.

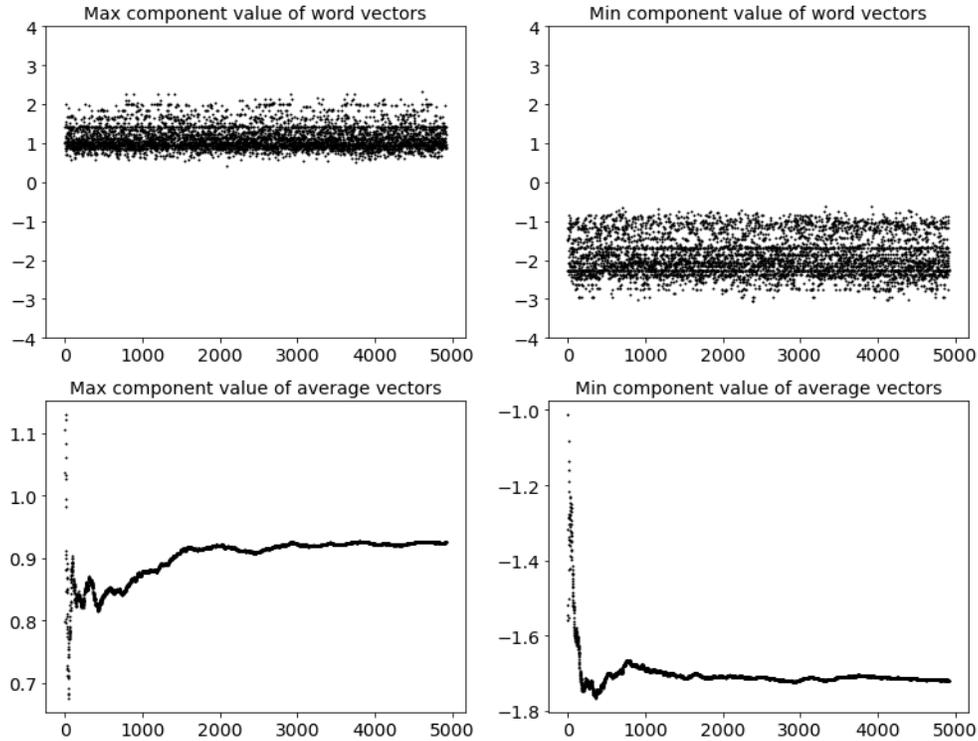

Figure A.3

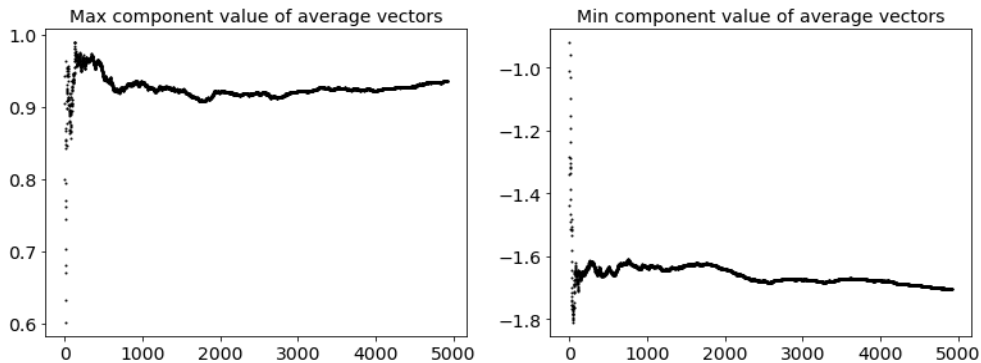

Figure A.4

In Figure A.5 we observe again the effect of averaging over the terms, but in this case considering all the components, not only the extreme ones. The top row represents Corpus 1 and the bottom Corpus 2. Again we observe the effect of convergence due to the anisotropy of the space of representations as we average over more vectors, and the large text similarity between the two corpora in the last column.

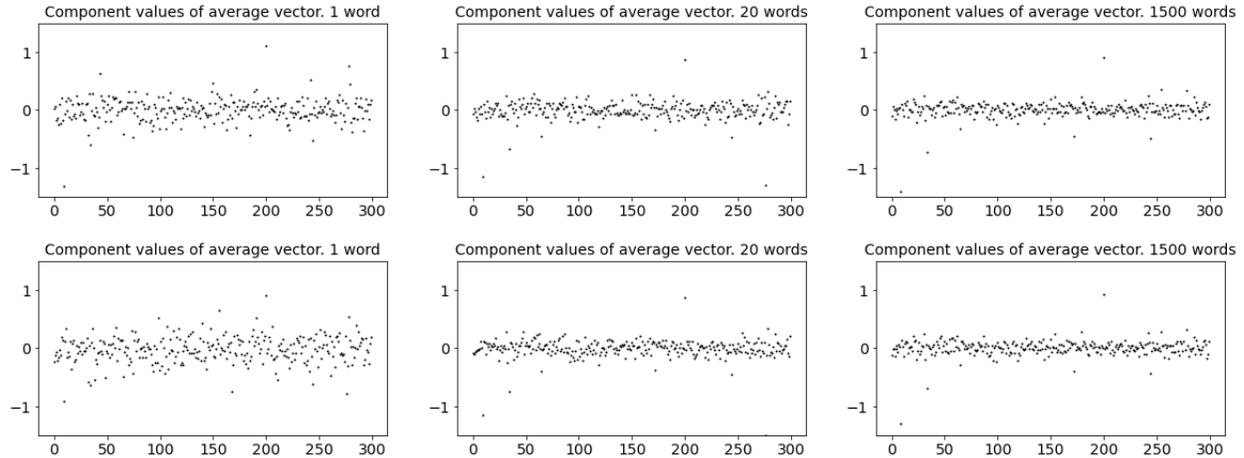

Figure A.5

In Figure A.6 we plot the cosine similarity between the average vectors of each corpus. For the term n we compute the average vector of the first n terms in each corpus, and then we compute the cosine similarity between both resulting vectors. Although the cosine similarity measure already performs a normalisation effect, by considering only the direction of the vectors, and not their magnitude, we see here how considering sufficiently long texts has a convergence effect with respect to the values of the cosine similarity. The range of possible values will therefore reduce and converge to the unit as the amount of text considered increases. This effect must be taken into account when assessing the differences between comparisons, and especially when comparing texts of different sizes.

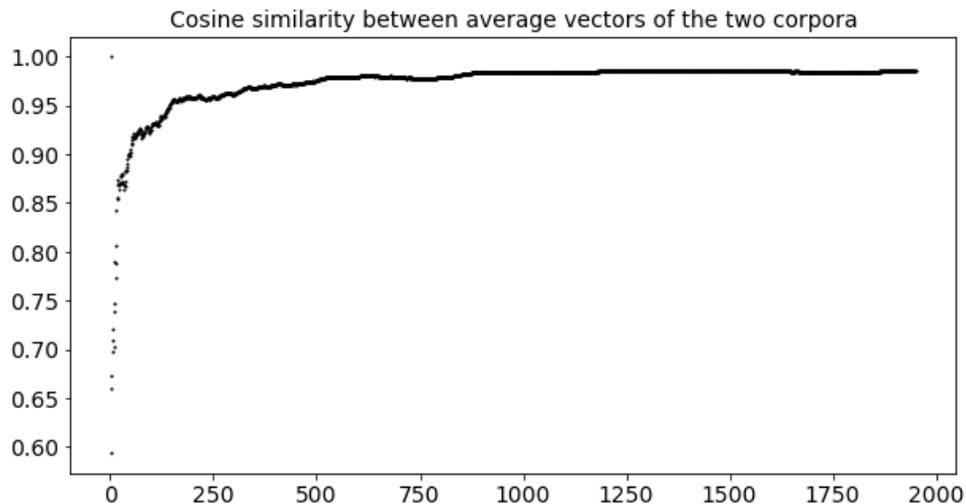

Figure A.6